\documentclass[conference]{IEEEtran}
\usepackage{times}

\usepackage[numbers]{natbib}
\usepackage{multicol}
\usepackage[bookmarks=true]{hyperref}

\usepackage{graphicx}
\usepackage{amsmath,amssymb}
\usepackage{booktabs}
\usepackage{siunitx}
\usepackage{subcaption}
\usepackage{algorithm}
\usepackage{algpseudocode}
\let\labelindent\relax
\usepackage{enumitem}
\usepackage{textcomp}
\usepackage{placeins}
\usepackage{multirow}
\usepackage{caption}
\usepackage{stfloats}

\title{\LARGE\bf A Principled Approach for Creating High-fidelity Synthetic Demonstrations for Imitation Learning}
\author{
\IEEEauthorblockN{\textbf{Moniruzzaman Akash}\hspace{2.5em}\textbf{Momotaz Begum}}
\IEEEauthorblockA{%
Department of Computer Science\\
University of New Hampshire\\
\{moniruzzaman.akash, momotaz.begum\}@unh.edu}
}

\begin{document}

\makeatletter
\let\@oldmaketitle\@maketitle
\renewcommand{\@maketitle}{\@oldmaketitle
  \vspace{-0.1em} 
  \begin{center}
    \begin{minipage}{\textwidth}
      \centering
      \includegraphics[width=\linewidth]{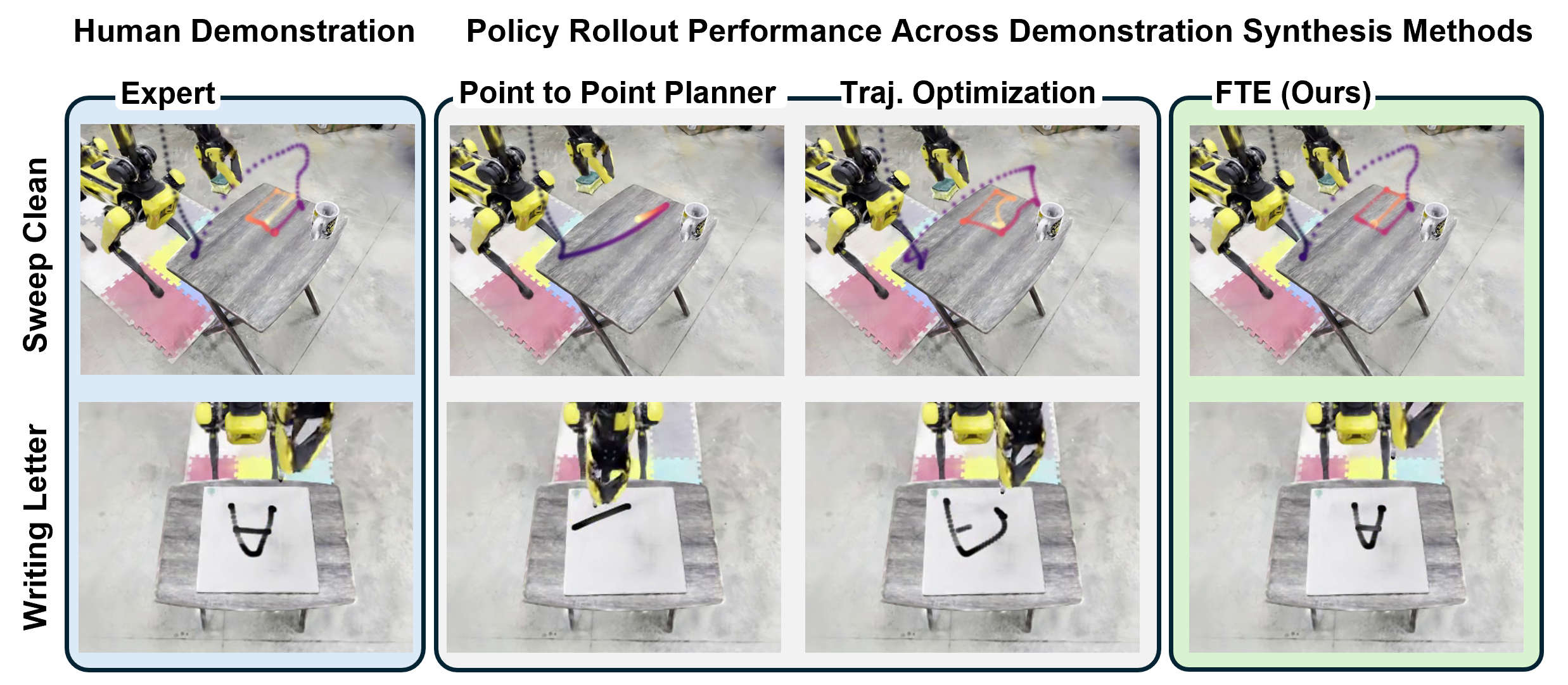}
      \captionof{figure}{Existing 3DGS-based demonstration generation synthesizes motion via point to point interpolated planning \cite{yang2025robosplat} or optimization \cite{pan20251001demos}, often distorting expert behavior. We propose Follow-the-Expert (\textbf{FTE}), a principled approach that synthesizes diverse yet shape and phase preserving trajectories with obstacle awareness}.
      \label{fig:overview}
    \end{minipage}
    \vspace{-1.9em}
  \end{center}
}

\maketitle

\begin{abstract}

Recent advances in 3D Gaussian Splatting (3DGS) have enabled visually realistic demonstration generation from a single expert trajectory and a short multi-view scan. However, existing 3DGS-based synthesis pipelines typically generate new motions using sampling-based planners or trajectory optimization, which often deviate substantially from the expert’s demonstrated path. While such deviations may be acceptable for tasks insensitive to motion shape, they discard subtle spatial and temporal structure that is critical for contact-rich and shape-sensitive manipulation, causing increased demonstration diversity to harm downstream policy learning.
We argue that demonstration synthesis should treat the expert trajectory as a strong prior. Building on this principle, we propose a framework that synthesizes diverse task demonstrations while explicitly preserving expert motion structure. We model the expert trajectory using Dynamic Movement Primitives (DMPs) and retarget it to new goals, object configurations, and viewpoints within a reconstructed 3DGS scene, yielding phase-consistent, shape-preserving motion by construction.
To safely realize this expert-preserving diversity in cluttered scenes, we introduce an analytic obstacle-aware DMP formulation that operates directly on the continuous density field induced by the 3DGS representation. This enables collision avoidance while minimally perturbing the nominal expert motion, unifying photorealistic rendering and geometric reasoning without additional scene representations.
We evaluate our approach on a Spot mobile manipulator across three manipulation tasks with increasing sensitivity to trajectory fidelity, using only a single human demonstration per task. Compared to planner- and optimization-based synthesis, our method produces trajectories with lower deviation and collision rates and yields higher task success when training diffusion-based visuomotor policies. Together, these results demonstrate that effective data augmentation for robot learning is not merely about maximizing diversity, but about generating expert-preserving diversity when task semantics are encoded in motion itself.
\end{abstract}


\section{Introduction}

Imitation learning enables robots to acquire complex manipulation skills from a set of human demonstrations \cite{Bagnell2015AnIT, zare2024survey}. However, visuomotor policies trained from limited data often overfit to the exact object poses, camera viewpoints, and scene configurations encountered during collection, leading to brittle behavior at deployment time. This challenge has motivated substantial recent work on demonstration synthesis and data augmentation, aiming to expand training distributions without the cost of additional real-world data collection \cite{torne2024reconciling}.

Recent pipelines based on \emph{3D Gaussian Splatting} (3DGS) \cite{kerbl2023gaussian} have made significant progress in this direction. From a single expert trajectory and a short multi-view scan, these methods reconstruct a photorealistic scene and render large numbers of novel demonstrations under diverse object poses, camera viewpoints, appearances, and lighting conditions~\cite{yang2025robosplat,pan20251001demos,xue2025demogen, yuan2025roboengine, chen2024rovi}. Policies trained on such data exhibit impressive robustness to visual distribution shift, highlighting the power of explicit 3D scene representations for data-efficient robot learning.

Despite these advances in visual generalization, existing 3DGS-based demonstration synthesis pipelines share a critical weakness: \emph{they do not preserve expert's wisdom encoded in the demonstrated trajectory}. In most prior work, new trajectories are generated by retargeting sparse keyframes and connecting them using sampling-based planners or trajectory optimization~\cite{yang2025robosplat,pan20251001demos}. While these methods produce geometrically feasible motions, they often deviate substantially from the demonstrated path, altering approach directions, temporal structure, and contact timing. Such deviations may be acceptable for tasks where only start and end poses matter, but they are detrimental for contact-rich and shape-sensitive manipulation, where success depends on subtle spatiotemporal structure encoded in the expert trajectory itself. Examples include sweeping a floor or a table top, making a specific pattern, or even simple pick-n-place tasks where a specific path is preferred to avoid coming closer to fragile or delicate objects. For such tasks, na\"{i}vely increasing the diversity of demonstrations can inject behavior that is visually plausible but semantically incorrect, ultimately harming downstream policy learning, as shown in Fig.~\ref{fig:overview}.

To address this issue, we propose FTE (Follow The Expert), a principled approach for synthesizing demonstrations while preserving the shape and phase of expert trajectory. FTE treats the expert trajectory as a strong prior.
We operationalize this principle by replacing planner-centric motion synthesis with \emph{Dynamic Movement Primitives} (DMPs)~\cite{ijspeert2013dmp} as a trajectory prior. By fitting a DMP to the expert end-effector path, we obtain a stable, phase-consistent dynamical system that captures the demonstrated trajectory's shape and timing. Demonstration synthesis then becomes a matter of retargeting the DMP to new goals and frames, rather than re-planning trajectories from scratch. This formulation preserves the expert’s spatiotemporal structure by construction, while still enabling systematic diversity over object poses and viewpoints.

To safely realize this expert-preserving diversity in cluttered real-world scenes, we introduce an analytic obstacle-aware DMP formulation that operates directly on the reconstructed 3DGS scene. We interpret the Gaussian parameters not only as a photorealistic radiance field, but also as a continuous density representation of scene geometry. Obstacle avoidance is enforced by smoothly modulating the DMP dynamics using gradients of this density field, guaranteeing minimum clearance while minimally perturbing the nominal expert motion. This unifies photorealistic rendering and continuous geometric reasoning within a single scene representation, without introducing additional simulators, point clouds, or signed distance fields.

We implemented the proposed FTE on the Boston Dynamics Spot mobile manipulator, collecting a single expert demonstration via VR teleoperation and synthesizing large demonstration sets across object poses and camera viewpoints. Across three manipulation tasks with increasing sensitivity to trajectory fidelity we show that FTE-based synthesis dramatically reduces trajectory deviation and collision rates compared to planner- and optimization-based baselines. Crucially, visuomotor policies trained on our synthesized data reproduce expert behavior more faithfully and achieve higher real-robot success when motion structure is essential for task success. 

Together, these results shed light on an interesting aspect of imitation learning  : effective data augmentation is not about maximizing diversity alone, but about generating diversity that preserves expert intent/wisdom when task semantics are encoded in the motion itself.

\textbf{Contributions.}
This paper makes the following contributions:
\begin{itemize}
    \item We introduce \emph{Follow-the-Expert (\textbf{FTE})} demonstration synthesis paradigm that treats the full expert trajectory as a motion prior, enabling phase-consistent, shape-preserving retargeting across object poses and viewpoints.
    \item We develop an analytic obstacle-avoidance formulation for DMPs that enforces clearance directly within a 3D Gaussian Splatting density field, unifying photorealistic rendering and continuous geometric reasoning.
    \item Through real-robot experiments, we demonstrate that expert-preserving synthesis yields safer trajectories and significantly improves downstream policy performance on tasks where motion structure is critical.
\end{itemize}

\section{Related Work}

\subsection{Imitation learning and data augmentation}
Learning manipulation from demonstrations is a longstanding route to skill acquisition~\cite{Argall2009Survey,Bagnell2015AnIT,zare2024survey}, but modern visuomotor policies remain data-hungry and can overfit to narrow initial-state and observation distributions. This has motivated broad efforts to improve robustness via synthetic data and sim2real—ranging from domain randomization and appearance perturbations~\cite{sadeghi2016cad2rl,tobin2017domainrand} to sim-to-sim adaptation for more data-efficient transfer~\cite{james2019sim}. In parallel, large-scale offline datasets and benchmarks have helped standardize evaluation and enabled training of high-capacity policies~\cite{robomimic2021,Chi2023DiffusionPolicy}. Our work targets a complementary bottleneck: when \emph{only one} or a few expert demonstrations are available, how can we expand coverage \emph{without corrupting the expert’s motion semantics}?

\subsection{Radiance fields and 3DGS for Real2Sim2Real and demo generation}
Neural radiance fields (NeRF) introduced a powerful paradigm for photorealistic view synthesis~\cite{Mildenhall2020}, and 3D Gaussian Splatting (3DGS) provides a computationally efficient explicit representation with real-time rendering~\cite{kerbl2023gaussian}. These advances have rapidly influenced Real2Sim2Real pipelines and robot learning augmentation: 3DGS-based simulators and renderers reduce the visual reality gap by reconstructing real scenes and producing photo-consistent training observations~\cite{qureshi2025splatsim,li2024robogsim,torne2024reconciling}. More recently, 3DGS has enabled \emph{one-shot demonstration generation} by editing the reconstructed scene (object pose, appearance, viewpoint, embodiment) and rendering large synthetic datasets from few real rollouts~\cite{yang2025robosplat,yuan2025roboengine}. These works establish 3DGS as a strong substrate for visual diversity. Our focus is on preserving the motion quality where we leverage 3DGS not only for rendering but also as the \emph{geometric field} against which obstacle-aware synthesis is performed.

\subsection{From feasibility-driven motion synthesis to expert-faithful retargeting}
Existing 3DGS-based demo generation pipelines commonly synthesize new trajectories via keyframes coupled with planning or trajectory optimization~\cite{yang2025robosplat,pan20251001demos,xue2025demogen}. This design is effective for producing feasible motions at scale, but it implicitly optimizes for task completion and collision-free execution, not for preserving the demonstrated spatiotemporal structure. As a result, the generated motion can drift in approach direction, contact timing, and phase progression—precisely the factors that matter in contact-rich or shape-sensitive behaviors, where the trajectory itself encodes task semantics. 
\par The proposed FTE treats the expert trajectory as a strong prior: Dynamic Movement Primitives (DMPs) provide a structured retargeting mechanism that preserves shape and phase by construction while still allowing goal/frame adaptation~\cite{ijspeert2013dmp,Pastor2009,Ude2014}. Our obstacle avoidance operates directly on the continuous density induced by 3DGS, aiming to introduce only the minimum deviation necessary for safety, rather than replanning from scratch. Table~\ref{tab:demo_comparison} summarizes how FTE differs from prior 3DGS-based augmentation and demo generation methods.

\begin{table*}[t]
\centering
\small
\setlength{\tabcolsep}{3.5pt}
\caption{Comparison of demonstration generation methods using 3D Gaussian Splatting. Motion generation highlights whether expert demonstrations constrain the synthesized trajectories.}
\label{tab:demo_comparison}
\begin{tabular}{l c c c c}
\toprule
\textbf{Method} &
\textbf{Scene Rep.} &
\textbf{Motion Generation} &
\textbf{Expert Path shape Preserved} &
\textbf{Obstacle Handling} \\
\midrule
SplatSim \cite{qureshi2025splatsim} &
3DGS + Sim &
Simulator expert \textit{(slow)} &
No Claim &
Physics engine \\
1001 DEMOS \cite{pan20251001demos} &
3DGS + PC &
Demo-anchored traj.\ opt. \textit{(slow)} &
Partial (pre-contact) &
TSDF / hull \\
RoboSplat \cite{yang2025robosplat} &
3DGS &
Keyframes + planner \textit{(med--slow)} &
No (goal only) &
No \\
DemoGen \cite{xue2025demogen} &
3DGS (render) &
Task-level traj.\ opt. \textit{(slow)} &
No &
Implicit (opt.) \\
RoboGSim \cite{li2024robogsim} &
3DGS + Sim &
Sim-generated trajs \textit{(slow)} &
No claim &
Physics engine \\
\textbf{FTE (Ours)} &
3DGS &
DMP prior \textit{(fast)} &
\textbf{Yes} &
3DGS density \\
\bottomrule
\end{tabular}
\vspace{-1em}
\end{table*}





\section{Method}
\label{sec:method}

\begin{figure*}[h!]
    \centering
    \includegraphics[width=1.0\linewidth]{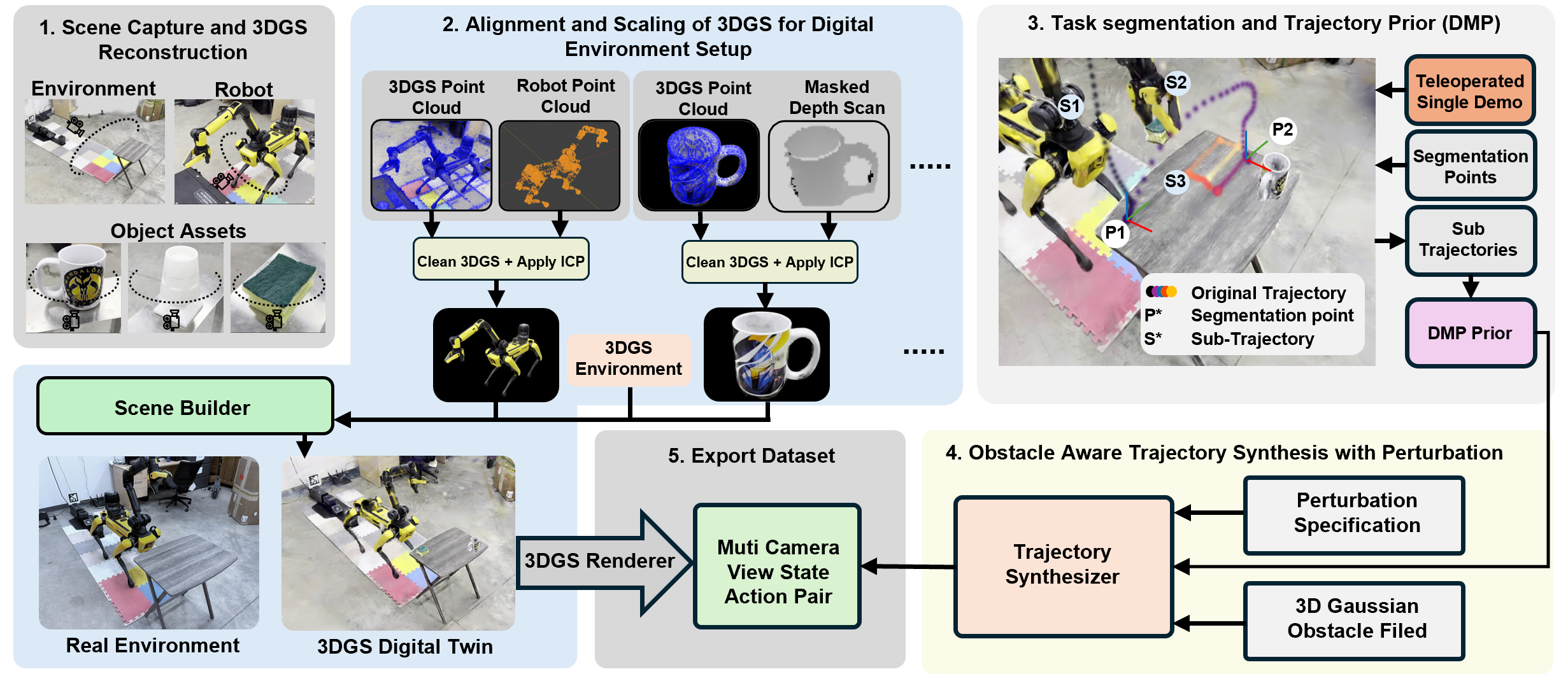}
    \caption{FTE pipeline. A single teleoperated demonstration is encoded as a DMP trajectory prior and retargeted within an aligned 3D Gaussian Splatting scene using density-based obstacle awareness, producing multi-view synthetic data for policy learning.}
    \label{fig:pipeline}
    \vspace{-1em}
\end{figure*}

Fig.~\ref{fig:pipeline} shows the pipeline for the proposed FTE. Inputs to the system are (i) one teleoperated expert demonstration, (ii) a one-shot scene scan reconstructed as a 3D Gaussian Splatting (3DGS) model, and (iii) user-specified perturbation distributions over task goals. 
Outputs are synthesized end-effector trajectories, photorealistic multi-view RGB observations rendered from 3DGS, and state--action datasets suitable for visuomotor policy learning.

\subsection{Scene Capture and 3DGS Reconstruction}
\label{sec:scene}

We reconstruct a static manipulation scene using a short handheld video sweep and
3D Gaussian Splatting (3DGS)~\cite{kerbl2023gaussian}.
Each Gaussian is parameterized by a mean $\boldsymbol{\mu}_i \in \mathbb{R}^3$,
covariance $\boldsymbol{\Sigma}_i$, opacity $\alpha_i$, and appearance coefficients.
The resulting representation provides a continuous, explicit model of the scene
that supports real-time photorealistic rendering from novel viewpoints.

Crucially, we use the reconstructed 3DGS not only as a visual renderer, but also as
a geometric substrate for motion reasoning.
The Gaussian density induced by the scene serves as a smooth proxy for occupied
space, enabling continuous evaluation of clearance and obstacle proximity during
trajectory synthesis.
To avoid spurious self-collisions, the robot is removed from the reconstruction
using segmentation masks during capture, yielding a scene-only Gaussian model.
This representation allows us to unify visual augmentation and geometric reasoning
within a single scene model, without introducing auxiliary simulators, point
clouds, or signed distance fields.

\subsection{Metric Alignment Between 3DGS and Robot Kinematics}
\label{sec:alignment}

To enable consistent reasoning between the reconstructed scene and the robot’s
motion, we align the 3DGS model to the robot’s kinematic frame.
Let $\mathcal{P}_{\text{scene}}$ denote a point set obtained by sampling Gaussian
centers $\{\boldsymbol{\mu}_i\}$ filtered by opacity, and let
$\mathcal{P}_{\text{robot}}$ denote a point-cloud proxy of the robot in a canonical
configuration derived from its kinematic model.

We estimate a rigid transformation
$T_{\text{gs}\rightarrow\text{robot}} = (\mathbf{R}, \mathbf{t}) \in SE(3)$
by solving a point-to-point Iterative Closest Point (ICP) problem~\cite{besl1992method}:
\begin{equation}
\min_{\mathbf{R}, \mathbf{t}} \sum_j \left\| \mathbf{R}\mathbf{p}_j + \mathbf{t} -
\mathbf{q}_j \right\|^2,
\quad \mathbf{p}_j \in \mathcal{P}_{\text{scene}},\;
\mathbf{q}_j \in \mathcal{P}_{\text{robot}} .
\end{equation}

The estimated transform is applied to all Gaussian means and orientations,
bringing the reconstructed scene into the robot’s coordinate frame.
We verify alignment by projecting the robot proxy into multiple camera views and visually confirming consistency with the reconstructed scene.
This alignment ensures that obstacle-aware trajectory modulation and expert
demonstration retargeting are performed in a shared metric space, which is
essential for preserving both motion structure and safety during synthesis.

\subsection{Interaction-Aware Expert Demonstration Segmentation}
\label{sec:segmentation}

A single expert demonstration typically interleaves distinct interaction phases
(e.g., free-space approach, contact manipulation, and withdrawal), and fitting a
single primitive over the full horizon can blur these regime changes.
We therefore segment each demonstration into $K$ temporally ordered subtasks at
salient interaction events, yielding split indices
$\mathcal{S}=\{t_0=1 < t_1 < \cdots < t_K=T\}$.
During teleoperation, the expert marks these boundaries online via a dedicated
controller button at interaction-mode transitions (e.g., pre-/post-grasp, contact
onset/offset, or kinematic mode switches), producing segment trajectories
\begin{equation}
\mathcal{T}_k = \{(\mathbf{x}_t,\mathbf{R}_t)\}_{t=t_{k-1}}^{t_k}, \quad k=1,\ldots,K,
\end{equation}
with boundary conditions at $(\mathbf{x}_{t_{k-1}},\mathbf{R}_{t_{k-1}})$ and
$(\mathbf{x}_{t_k},\mathbf{R}_{t_k})$.

Event-aligned segmentation improves DMP fitting by isolating approximately
single-mode dynamics per segment and makes goal retargeting well-posed by
perturbing only the appropriate terminal conditions.
Task segmentation has been studied extensively, including learning skill
boundaries and task graphs from unstructured demonstrations~\cite{Konidaris2012CST,Niekum2015FSM}
and latent option discovery~\cite{Krishnan2017DDCO}.
However, these approaches often require multiple demonstrations and can be brittle
under contact noise, introducing boundary jitter that degrades phase-consistent
reproduction~\cite{Eiband2023OnlineSeg}.
In our setting, subtasks correspond to discrete interaction events, so operator
event marking provides a simple, reproducible, and human-interpretable
decomposition that directly supports expert-faithful fitting and retargeting.

\subsection{Dynamic Movement Primitive Prior}
Given the interaction-aware segments $\{\mathcal{T}_k\}_{k=1}^{K}$ from
Sec.~\ref{sec:segmentation}, we model each segment with Dynamic Movement
Primitives (DMPs)~\cite{ijspeert2002movement, schaal2006learning, ijspeert2013dmp}.
DMPs provide a stable dynamical system with a learned nonlinear forcing term,
allowing smooth reproduction while adapting to new boundary conditions.
This property is central to our synthesis pipeline: in Sec.~\ref{sec:perturb} we
generate diversity by perturbing segment goals while keeping the learned forcing
term fixed, thereby preserving the expert’s characteristic motion shape and
phase progression.

\paragraph{Position DMP.}
For a 1D trajectory $y(t)$, the transformation system, defined as 
\begin{align}
\tau \dot{v}(t) &= \alpha_z\big(\beta_z(g - y(t)) - v(t)\big) + f(s(t)), \\
\tau \dot{y}(t) &= v(t),
\end{align}
is driven by the following canonical phase variable
\begin{equation}
\dot{s}(t) = -\alpha_s s(t), \qquad s(0)=1.
\end{equation}
We represent the forcing term using normalized radial basis functions (RBFs),
\begin{equation}
f(s) = \frac{\sum_{i} w_i \psi_i(s)}{\sum_{i} \psi_i(s)}\, s\,(g-y_0),
\qquad
\psi_i(s)=\exp\!\big(-h_i(s-c_i)^2\big),
\end{equation}
and learn weights $\{w_i\}$ by ridge regression so that the resulting accelerations
match the demonstrated segment (after time normalization by $\tau$).
For Cartesian trajectories, we fit one DMP per axis and roll them out jointly to
reproduce the 3D end-effector path.

\paragraph{Orientation DMP.}
To avoid discontinuities from component-wise quaternion regression, we model
orientation in a local, continuous chart using the quaternion log map.
Let $q_t$ be the demonstrated orientation and $q_0$ the segment start.
We form relative orientations and map them to rotation vectors:
\begin{equation}
r_t = \log\!\left(q_0^{-1} \otimes q_t\right) \in \mathbb{R}^3 .
\end{equation}
We unwrap the rotation-vector sequence $\{r_t\}$ to maintain continuity and fit a
3D DMP to $r(t)$.
At reproduction time, we recover the orientation as
\begin{equation}
q(t) = q_0 \otimes \exp\!\big(\hat{r}(t)\big).
\end{equation}
This representation yields smooth rotational reproduction and supports goal
retargeting in Sec.~\ref{sec:perturb} by modifying the segment terminal pose while
preserving the expert’s rotational structure.

\subsection{Goal Perturbation for Demonstration Augmentation}
\label{sec:perturb}

\begin{figure}[h]
    \centering
    \includegraphics[width=\linewidth]{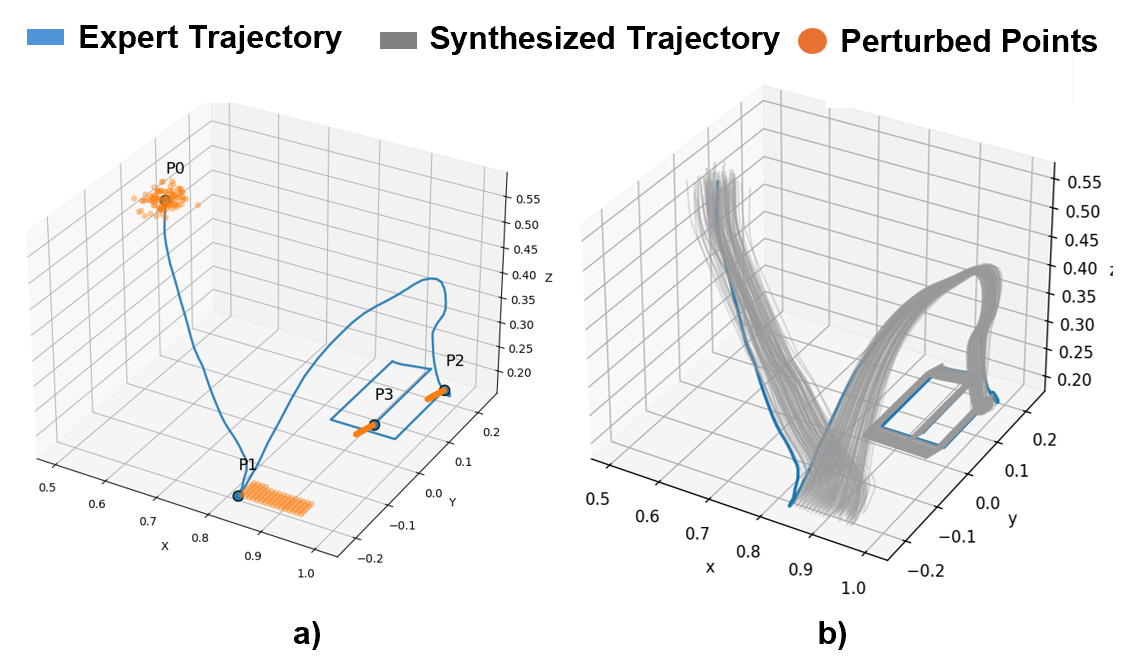}
    \caption{(a) Perturbed boundary points $P_0$--$P_3$ (orange) on the expert path (blue). (b) Synthesized DMP rollouts (gray) retargeted to these perturbations.}
    \label{fig:perturbation_and_synth}
    \vspace{-1em}
\end{figure}

To generate diverse demonstrations, we perturb the boundary poses of each segment while keeping the learned forcing term fixed.
For a segment with start $y_0$ and goal $g$, we sample
\begin{equation}
\tilde{g} = g + \Delta p, \quad
\tilde{q} = q \otimes \Delta q,
\end{equation}
where $\Delta p$ and $\Delta q$ are drawn from user-specified bounded gaussian distributions over translation and rotation.
The DMP then reproduces a trajectory connecting $(y_0,q_0)$ to $(\tilde{g},\tilde{q})$, preserving phase and shape. Fig.~\ref{fig:perturbation_and_synth} illustrates the boundary perturbation strategy and the resulting synthesized rollouts for a representative segmented trajectory.

\subsection{Density-Based Obstacle Avoidance in 3DGS}
\label{sec:obstacle}

Prior demonstration-generation pipelines often introduce an additional geometric
proxy (e.g., point clouds, meshes, or separate simulation/SDF models) for collision
checking, increasing pipeline complexity and requiring extra reconstruction.
In contrast, we reuse the same aligned 3DGS representation for both rendering and
safety reasoning, extracting obstacle cues directly from the density field without
any auxiliary scene model.
This aligns with recent 3DGS-based safety work such as Splat-Nav~\cite{chen2024splat}, but
we use density reasoning to \emph{locally modulate} expert-faithful DMP rollouts for
demonstration synthesis rather than global navigation planning.

\begin{figure}[h]
    \centering
    \includegraphics[width=1.0\linewidth]{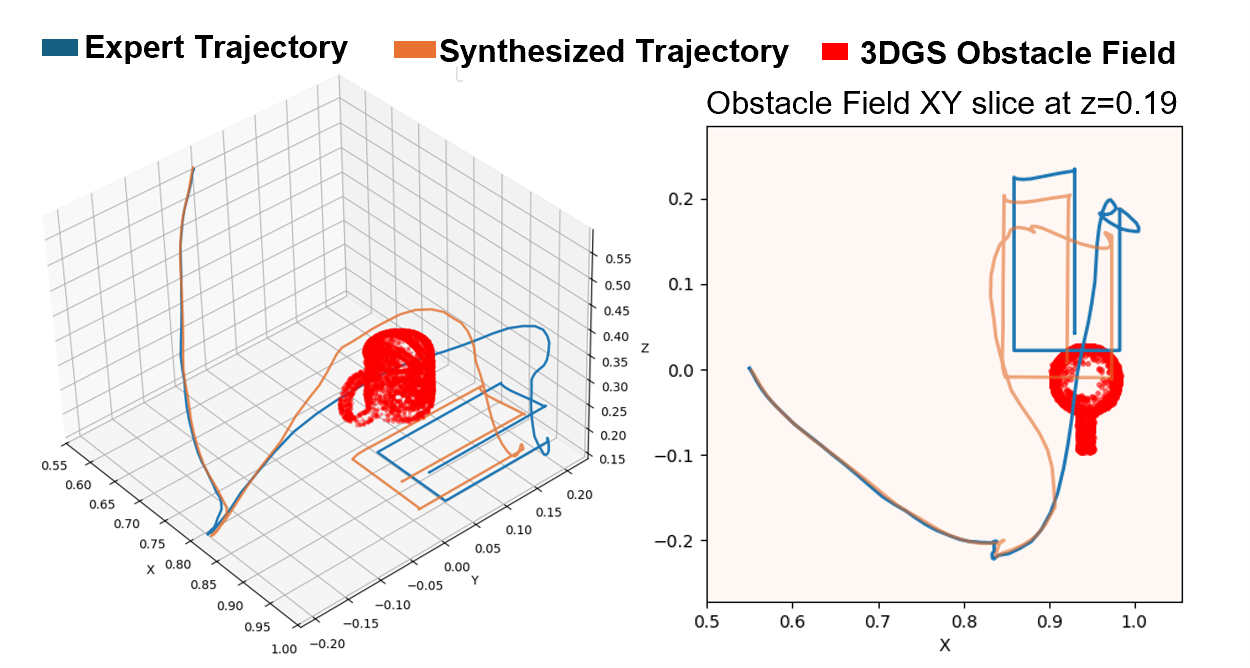}
    \caption{Obstacle-aware retargeting using a 3DGS density field. A density-gradient coupling locally deforms the rollout around dense regions while preserving the expert motion structure.}
    \label{fig:obstacle_avoidance}
    \vspace{-1em}
\end{figure}

The aligned 3D Gaussian Splatting (3DGS) map induces a continuous density field
that we use as a proxy for scene occupancy (Fig.~\ref{fig:obstacle_avoidance}).
At any location $x\in\mathbb{R}^3$, we define
\begin{equation}
\rho(x) = \sum_{i} \alpha_i
\exp\!\left(-\tfrac{1}{2}(x-\mu_i)^\top \Sigma_i^{-1}(x-\mu_i)\right),
\label{eq:gs_density}
\end{equation}
where $(\mu_i,\Sigma_i,\alpha_i)$ are the parameters of the $i$-th Gaussian.
We compute $\rho(x)$ from the full 3DGS field, but evaluate it efficiently via local neighborhood queries, which preserve smoothness in practice.

During DMP reproduction, we monitor $\rho(x)$ along the end-effector trajectory
and activate obstacle avoidance when $\rho(x)>\rho_{\mathrm{th}}$ by injecting a
coupling term into the DMP transformation dynamics (cf.~DMP obstacle
couplings~\cite{Park2008Humanoids,ijspeert2013dmp}).
We estimate $\nabla\rho(x)$ numerically (central differences) and define an
outward normal direction
\begin{equation}
\hat n(x) = -\frac{\nabla \rho(x)}{\|\nabla \rho(x)\|+\varepsilon}.
\label{eq:normal}
\end{equation}
Let $\hat v = v/(\|v\|+\varepsilon)$ denote the normalized instantaneous velocity.
Rather than choosing an arbitrary tangential direction, we use the component of
$\hat n$ orthogonal to the motion direction, i.e.,
$\tilde t(x)=\hat n(x)-(\hat n(x)^\top \hat v)\hat v$ and
$\hat t(x)=\tilde t(x)/(\|\tilde t(x)\|+\varepsilon)$, which encourages sliding
around dense regions without directly opposing forward progress.
The resulting obstacle-induced acceleration is
\begin{equation}
a_{\mathrm{obs}}(x,v) =
\lambda\!\big(\rho(x),\,\hat v^\top \hat n(x)\big)\,
\Big(\hat n(x) + \gamma\,\hat t(x)\Big),
\label{eq:obs_accel}
\end{equation}
where $\gamma$ controls tangential bias.
The gain $\lambda(\cdot)$ gates the response using both density magnitude
(relative to $\rho_{\mathrm{th}}$) and whether the motion is directed into the
obstacle (via $\hat v^\top \hat n$); in addition, we evaluate density with a short
lookahead along $\hat v$ to react earlier to impending collisions.
We add this term directly to the DMP acceleration,
\begin{equation}
\tau \dot v = a_{\mathrm{dmp}} + a_{\mathrm{obs}},
\label{eq:dmp_coupling}
\end{equation}
so the trajectory locally deforms around obstacles while the DMP prior preserves
the global expert motion structure.

To prevent drift after clearing obstacles, we include a bounded \emph{return-to-reference} correction that gently pulls the rollout toward the nominal phase-indexed DMP trajectory once density decreases; this correction is automatically down-weighted near high-density regions to avoid fighting the repulsive term and improves consistency across augmented rollouts without iterative replanning.

\begin{algorithm}[t]
\caption{Expert-Faithful Demonstration Synthesis (FTE) with 3DGS Obstacle Avoidance}
\label{alg:dmp_synthesis}
\begin{algorithmic}[1]
\Require Expert trajectory $\mathcal{T}=\{(\mathbf{x}_t,\mathbf{R}_t)\}_{t=1}^{T}$, split indices $\mathcal{S}$,
perturbation specs $\{\Delta_k\}$, 3DGS Gaussians $\{\boldsymbol{\mu}_i,\boldsymbol{\Sigma}_i,\alpha_i\}_{i=1}^{N}$,
threshold $\rho_{\text{th}}$, gain $\lambda$
\Ensure Synthesized trajectory $\hat{\mathcal{T}}$

\State Decompose $\mathcal{T}$ into segments $\{\mathcal{T}_k\}_{k=1}^{K}$ using $\mathcal{S}$
\For{$k \leftarrow 1 \dots K$}
    \State \textbf{Fit DMP:} estimate phase $s(t)$; fit forcing term weights for $\mathbf{x}(t)$ and orientation
    \State \textbf{Retarget goal:} sample terminal perturbation $\delta_k \sim \Delta_k$ and set $g_k' = g_k \oplus \delta_k$
    \State Initialize state $(\mathbf{y},\mathbf{v}) \leftarrow (\mathbf{x}^{(k)}_{0}, \mathbf{0})$
    \For{$n \leftarrow 1 \dots N_{\text{roll}}$}
        \State Update phase $s \leftarrow s - \alpha_s s \Delta t$
        \State DMP acceleration $\mathbf{a}_{\text{dmp}} \leftarrow \alpha_z(\beta_z(g_k' - \mathbf{y}) - \mathbf{v}) + f(s)$
        \State Obstacle term $\mathbf{a}_{\text{obs}} \leftarrow \textsc{3DGSRepel}(\mathbf{y}; \rho_{\text{th}}, \lambda)$
        \State Integrate $\mathbf{v} \leftarrow \mathbf{v} + (\mathbf{a}_{\text{dmp}} + \mathbf{a}_{\text{obs}})\Delta t$
        \State \hspace{1.6em} $\mathbf{y} \leftarrow \mathbf{y} + \mathbf{v}\Delta t$
        \State Append $\mathbf{y}$ to synthesized segment $\hat{\mathcal{T}}_k$
    \EndFor
\EndFor
\State Concatenate segments $\hat{\mathcal{T}} \leftarrow \hat{\mathcal{T}}_1 \Vert \cdots \Vert \hat{\mathcal{T}}_K$
\State \Return $\hat{\mathcal{T}}$
\end{algorithmic}
\end{algorithm}
\vspace{-0.5em}

\subsection{Demonstration Synthesis and Dataset Export}
\label{sec:synthesis}

For each subtask, we generate multiple augmented trajectories by combining:
(i) terminal pose perturbations,
(ii) DMP rollout with obstacle-aware modulation, and
(iii) consistent rendering from the aligned 3DGS scene.

Augmented trajectories are concatenated across subtasks to form complete task demonstrations. This process is computationally efficient, as DMP rollout is closed-form and does not require iterative optimization making synthesis computationally efficient.
Algorithm~\ref{alg:dmp_synthesis} summarizes the complete demonstration synthesis procedure, integrating segment-wise DMP fitting, goal perturbation, and density-based obstacle avoidance.

\vspace{-2.0em}
\section{Experimental Setup}

We evaluate the proposed FTE framework on three manipulation tasks. 
The experiments are designed to answer the following questions:

\begin{enumerate}[label=\alph*), noitemsep, topsep=0pt]
    \item How does preserving the shape of expert trajectory during demonstration synthesis improve downstream policy performance?
    \item How does DMP-based synthesis compare to trajectory optimization and planner-based stitching used in prior work? \cite{yang2025robosplat} \cite{pan20251001demos}
    \item How does obstacle-aware modulation in a unified 3DGS representation improve safety without degrading motion fidelity?
\end{enumerate}

All experiments are conducted on a Spot mobile manipulator with a 7-DoF arm in real-world settings with a 3DGS digital twin renderer. Meta Quest controllers were used to collect demonstration by logging synchronized robot states, images, gripper, and segmentation events.

\subsection{Tasks}

\begin{figure}[h!]
    \centering
    \includegraphics[width=1.0\linewidth]{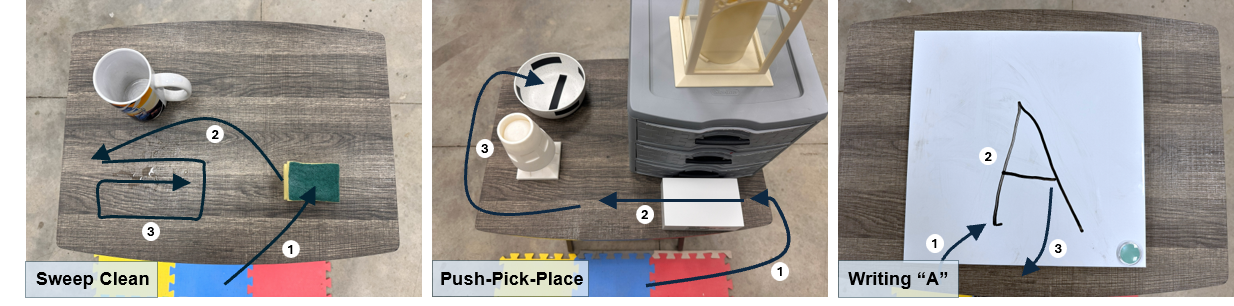}
    \caption{We choose 3 tasks where the trajectory encodes semantics and context: Sweep Clean, Push-Pick-Place, and Writing a letter "A".}
    \label{fig:task_description}
    \vspace{-1em}
\end{figure}
We consider three tasks, each collected from a single human demonstration, illustrated on Fig.~\ref{fig:task_description}:

\textbf{Sweep Cleaning.}
The robot grasps a sponge and wipes a coffee spill near a cup used as a visual cue.
Success requires continuous contact and stable end-effector orientation, making the
task sensitive to trajectory-shape distortion; deviations can lead to ineffective
cleaning or unsafe contact.

\textbf{Push--Pick--Place.}
The robot pushes a box from a cluttered region to an accessible area, then grasps
and places it into a bowl. The push is trajectory-sensitive—contact point, direction,
and path determine the final box pose—while the overall task also requires obstacle
avoidance and precise grasping.

\textbf{Letter Writing.}
The robot writes ``A'' on a planar surface with a marker. Performance depends on
spatiotemporal fidelity of the demonstrated stroke: even small deviations produce
visible errors. Unlike simple pick-and-place, success is largely determined by
reproducing the expert’s continuous motion profile, making this a stringent test of
the motion prior.

\subsection{Baselines}
We compare our approach against two closely related motion synthesis strategies that represent common alternatives to trajectory priors.

\textbf{Planner-Based Stitching (MPLib).}
This baseline follows the motion synthesis pipeline used in RoboSplat~\cite{yang2025robosplat}.
Expert demonstrations are retargeted at the keyframe level, and a sampling-based planner (MPLib) generates feasible trajectories between keyframes. It does not support obstacle avoidance.

\textbf{Demo-Anchored Trajectory Optimization (TrajOpt).}
This baseline uses the trajectory-optimization action generator described in 1001~DEMOS~\cite{pan20251001demos}. In 1001~DEMOS, optimization is applied primarily to the \emph{approach (pre-contact)} portion of the demonstration to produce smooth, collision-aware motions that remain close to the demonstrated approach while reaching a specified terminal pose. Following this design, we treat the expert approach segment as a soft anchor and synthesize new approach trajectories for our perturbed terminal poses. Since an official implementation is not released at the time of submission, we re-implement this TrajOpt baseline from the loss objective terms specified in the paper (e.g., demonstration funnel/anchoring, collision avoidance, smoothness, and view-friendly constraints), and use it solely as the motion generator; all other pipeline stages exactly follow our method.

\subsection{Evaluation Protocol}
\label{sec:eval_protocol}
For each task and synthesis method, we generate 256 augmented demonstrations from a
single expert trajectory and train a visuomotor Diffusion Policy~\cite{Chi2023DiffusionPolicy}
using identical architectures, observation modalities, and hyperparameters across
all methods.
Training datasets consist of the original human demonstration together with the
corresponding synthesized trajectories.
Each trained policy is evaluated over 40 rollout episodes per task under object
configurations that differ from the human demonstration scene.
For the \emph{Sponge Clean} and \emph{Push--Pick--Place} tasks, we additionally
evaluate methods with obstacle-aware synthesis (ours and TrajOpt) in
obstacle-augmented scenes, while methods without obstacle handling are evaluated
only in obstacle-free settings.
For the \emph{Writing} task, no obstacles are introduced, as the task is primarily chosen to test expert trajectory fidelity rather than environmental interaction.

\subsection{Metrics}
\label{sec:metrics}
We evaluate both task-level performance and trajectory-level fidelity using the following metrics.

\textbf{Task Success Rate.}
A rollout is considered successful if the task-specific objective is completed (e.g., spill fully wiped, object placed into the target receptacle, or letter written legibly).
For real-robot experiments, we report success rate over 40 trials per condition.

\textbf{Path Deviation (DTW).}
To measure adherence to the demonstrated motion, we compute Dynamic Time Warping
(DTW)~\cite{sakoe1978dynamic} between the expert trajectory and each synthesized
or executed trajectory after time normalization, reporting mean$\pm$std.
DTW is evaluated separately for end-effector position and orientation.
In general, for sequences $\mathbf{a}_{1:T}$ and $\mathbf{b}_{1:T^\ast}$,
\begin{equation}
\mathrm{DTW}(\mathbf{a},\mathbf{b}) =
\min_{\pi} \sum_{(i,j)\in\pi} d(\mathbf{a}_i,\mathbf{b}_j),
\end{equation}
where $\pi$ is a monotone warping path and $d(\cdot,\cdot)$ is the per-step
distance metric.
Lower DTW indicates stronger preservation of the expert trajectory shape and
orientation profile.



\textbf{Collision Rate.}
A collision is recorded if the end-effector enters a high-density region of the 3DGS scene, i.e.,
$\rho(\hat{\mathbf{x}}_t) > \rho_{\text{th}}$ for any $t$.
We report the percentage of rollouts containing at least one such event.

\textbf{Writing Error.}
For the letter-writing task, we evaluate spatial accuracy by rasterizing both the
expert and executed end-effector trajectories into binary images,
$I_{\text{exp}}$ and $I_{\text{exec}}$.
To ensure comparability, we first crop the whiteboard region from each image and
scale the trajectories to a common bounding box at a fixed resolution and stroke
width.
Writing error is measured as the normalized $\ell_1$ pixel difference,
\begin{equation}
E_{\text{write}} =
\frac{\lVert I_{\text{exec}} - I_{\text{exp}} \rVert_1}{\lVert I_{\text{exp}} \rVert_1},
\end{equation}
which reports the fraction of mismatched pixels relative to the expert trace.
Lower values indicate more faithful reproduction of the demonstrated character
shape.


\section{Results}
\label{sec:results}

\begin{figure*}[h!]
    \centering
    \includegraphics[width=0.82\linewidth]{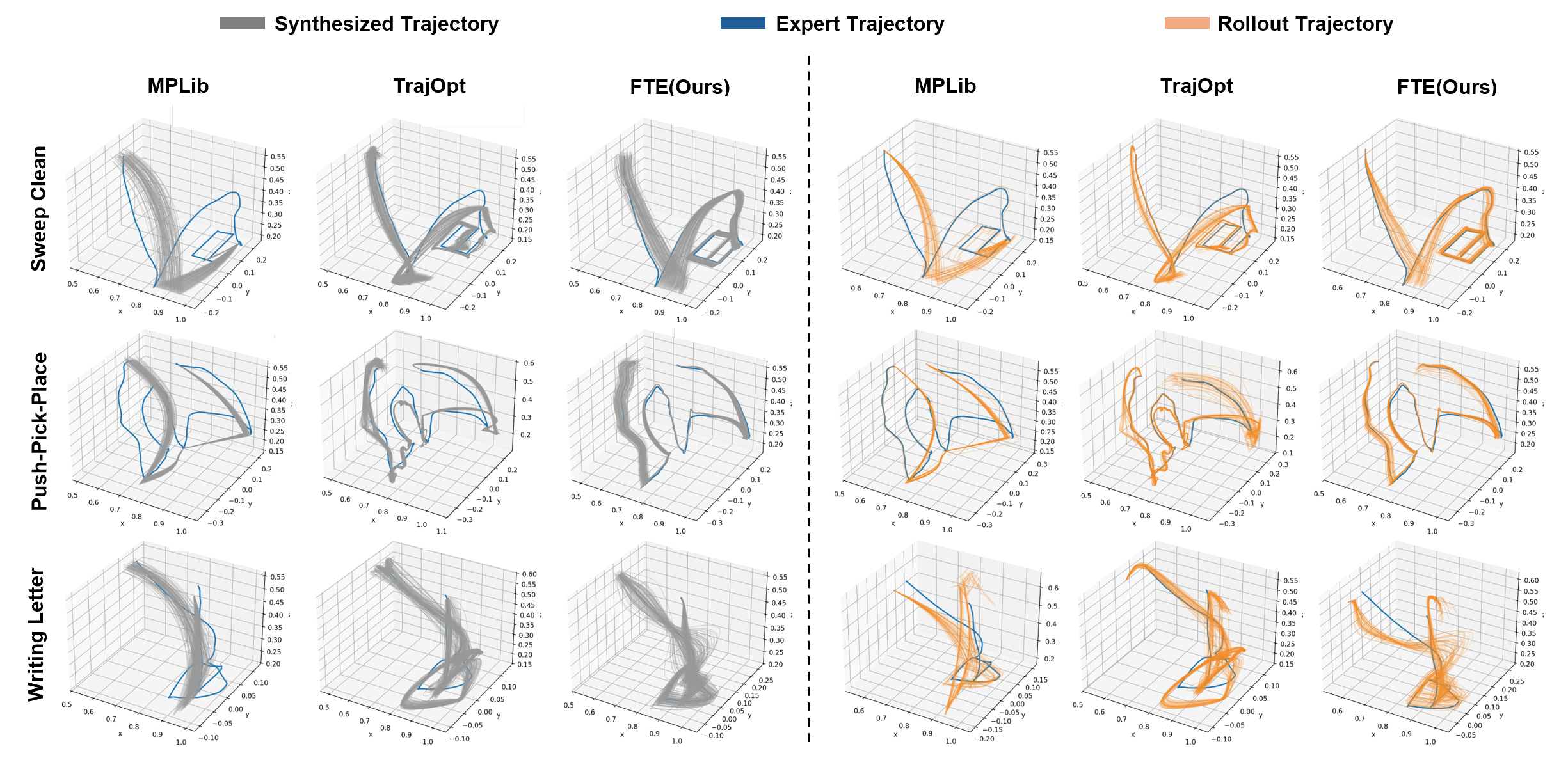}
    \caption{Synthesized and Rollout Trajectories across tasks.}
    \label{fig:rollout_paths}
    \vspace{-1em}
\end{figure*}

We evaluate \emph{Follow-the-Expert (FTE)} along three axes that mirror the paper’s claims: (i) whether synthesized demonstrations preserve expert motion structure,
(ii) whether this fidelity transfers to \emph{executed} policy rollouts, and
(iii) whether obstacle-aware synthesis improves safety without erasing the expert prior. Unless noted otherwise, all reported numbers are averaged over the rollout evaluation protocol in Sec.~\ref{sec:eval_protocol}.

\subsection{Trajectory Fidelity Transfers from Synthesis to Policy Execution}
\label{sec:results_fidelity}

A key question is whether ``expert-preserving'' synthesis is measurable at the
trajectory level \emph{and} whether it persists after policy learning.
We compute DTW~\cite{sakoe1978dynamic} to the expert demonstration for (i) the
synthesized trajectories and (ii) the trajectories executed by policies trained
on those synthesized datasets.
DTW is reported separately for end-effector position and orientation; for
orientation we use the geodesic distance on unit quaternions
($2\arccos(|q_1^\top q_2|)$)~\cite{huynh2009metrics}, accumulated along the DTW
alignment path and normalized by the path length.

\begin{figure}[h!]
    \centering
    \includegraphics[width=1.0\linewidth]{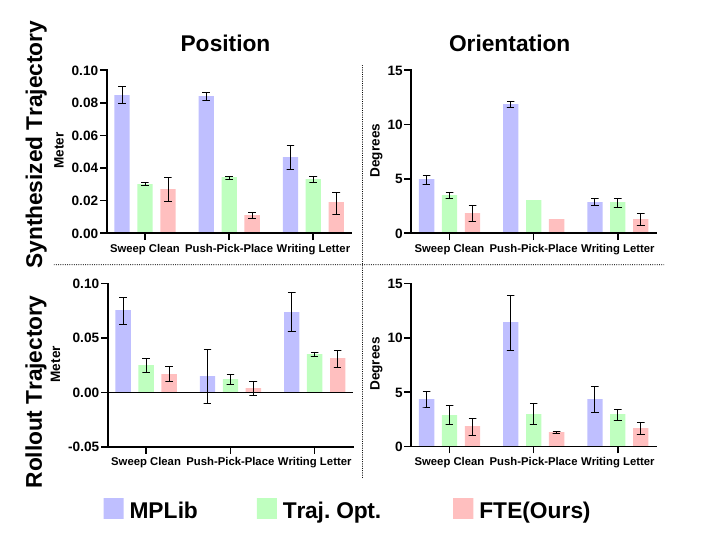}
    \caption{Dynamic Time Warping with respect to Expert Demo of Rolled out Trajectories across tasks. ($\downarrow$ Lower is better)}
    \label{fig:dtw_comparison}
    \vspace{-1.9em}
\end{figure}

Across all three tasks, FTE achieves the lowest DTW at the synthesis stage,
indicating that retargeting with a trajectory prior better preserves the expert’s
spatiotemporal structure than planner-based stitching or demo-anchored trajectory
optimization (Fig.~\ref{fig:dtw_comparison}).
This ordering persists at execution time: policies trained on FTE data produce
rollouts closer to the expert in both position and orientation, suggesting that
the synthesis prior shapes the behavior of the learned policy.
The effect is strongest in writing, where small phase or stroke deviations lead
to visible distortions; Fig.~\ref{fig:rollout_paths} shows that FTE preserves the
intended stroke structure rather than merely reaching similar endpoints, while
baselines exhibit larger shape distortions in curved and contact-sensitive
segments.

\subsection{Safety and Task Success Under Obstacle-Aware Synthesis}
\label{sec:results_safety_success}

\begin{table}[t]
\centering
\scriptsize
\setlength{\tabcolsep}{7pt}
\caption{Task success and safety across policy rollouts (mean over $N{=}40$ executions per task).
Collision rate is measured only for methods that perform obstacle-aware synthesis (FTE+OA and TrajOpt);
OA denotes obstacle-aware FTE rollouts.}
\label{tab:traj_fidelity}
\begin{tabular}{l l c c}
\toprule
\textbf{Task} & \textbf{Method} &
\textbf{Coll.\ (\%)$\downarrow$} &
\textbf{Succ.\ (\%)$\uparrow$} \\
\midrule
\multirow{4}{*}{Sweeping}
& MPLib (RoboSplat)    & N/A & 12.5 \\
& TrajOpt (1001-demo) & 30.5  & 75.0 \\
& FTE (ours)          & N/A & \textbf{92.5} \\
& FTE+OA (ours)       & \textbf{10.0} & 80.0 \\
\midrule
\multirow{4}{*}{Push--pick--place}
& MPLib (RoboSplat)    & N/A & 2.5 \\
& TrajOpt (1001-demo) & 67.5  & 30.0 \\
& FTE (ours)          & N/A & \textbf{75.0} \\
& FTE+OA (ours)       & \textbf{20.0} & 50.0 \\
\midrule
\multirow{4}{*}{Write ``A''}
& MPLib (RoboSplat)    & N/A & 0.0 \\
& TrajOpt (1001-demo) & N/A & 5.0 \\
& FTE (ours)          & N/A & \textbf{62.5} \\
\bottomrule
\end{tabular}
\vspace{-1.9em}
\end{table}

Next, we evaluate task success and safety during policy execution.
In obstacle-free settings, policies trained on FTE demonstrations achieve the
highest success rates on the two contact-sensitive manipulation tasks as well as
the writing task, consistent with the fidelity results above: when task semantics
are encoded in the motion profile (contact direction, approach geometry, timing),
preserving the expert’s structure yields more reliable behavior at deployment.

For cluttered scenes, we additionally evaluate obstacle-aware synthesis.
FTE+OA reduces collision rate substantially relative to the optimization baseline,
while maintaining competitive success. This highlights a practical trade-off:
purely expert-following synthesis maximizes motion fidelity, whereas adding OA
introduces only the \emph{minimal} deformation needed for clearance, improving
safety without reverting to global replanning. Importantly, the OA coupling is
computed directly from the aligned 3DGS density field, reusing the same scene
representation for rendering and collision reasoning rather than introducing a
separate geometric proxy (Sec.~\ref{sec:obstacle}).

\subsection{Writing Quality via Normalized Pixel Error}
\label{sec:results_writing}

Finally, we isolate shape-sensitive behavior by evaluating the writing task with
a normalized pixel error metric computed from rasterized stroke images (Sec.~\ref{sec:metrics}).

\begin{table}[h]
\centering
\scriptsize
\setlength{\tabcolsep}{7pt}
\caption{Normalized writing error for the letter-writing task (mean $\pm$ std over $N{=}40$ rollouts).
Lower values indicate more faithful reproduction of the expert-written character.}
\label{tab:writing_error}
\begin{tabular}{l c}
\toprule
\textbf{Method} & \textbf{Norm. Writing Error$\downarrow$} \\
\midrule
MPLib (RoboSplat)    & $0.170 \pm 0.033$ \\
TrajOpt (1001-demo) & $0.129 \pm 0.017$ \\
FTE (ours)          & $\mathbf{0.078 \pm 0.028}$ \\
\bottomrule
\end{tabular}
\vspace{-1.9em}
\end{table}

FTE achieves the lowest normalized writing error, indicating that the executed
letters match the expert’s stroke geometry more closely than either planner-based
or optimization-based synthesis.
Combined with the DTW findings, this result supports a core takeaway: for tasks
where the trajectory itself is the task specification, enforcing
expert-preserving structure during data synthesis is not merely an aesthetic
preference it directly improves the quality of learned policy executions.

\subsection{Takeaway and Limitation}
\label{sec:results_takeaway}
Across tasks spanning contact-rich wiping, non-prehensile pushing, and
shape-sensitive writing, the results support our hypothesis: demonstration
augmentation is most effective when it increases coverage \emph{without}
discarding the expert’s spatiotemporal ``wisdom.''
FTE preserves expert motion structure during synthesis, this fidelity transfers
to diffusion-based visuomotor policies~\cite{Chi2023DiffusionPolicy}, and
obstacle-aware synthesis improves safety using the same 3DGS representation
already required for photorealistic rendering.
A limitation is that obstacle reasoning relies on a conservative 3DGS density
proxy that can yield false positives under imperfect reconstructions; moreover,
we assume static scenes and do not explicitly model contact forces, so performance may degrade in highly dynamic environments or tasks requiring precise
force/impedance control.
Extending the approach to dynamic scenes and deformable objects is a promising
direction for future work.

\section{Conclusion}
We introduced a follow-the-expert synthesis paradigm that pairs 3DGS visual augmentation with DMP-based motion generation. By preserving expert path shape, timing and adding analytic obstacle avoidance we produce safer, more faithful rollouts in trajectory sensitive tasks that translate into higher policy success on a mobile manipulator. The approach is simple to adopt and complements existing 3DGS pipelines. Future work includes hybrid contact DMPs, uncertainty-aware clearance, and closed-loop replanning with learned residuals.




\bibliographystyle{plainnat}
\bibliography{references}


\end{document}